\documentclass{article}
\usepackage{spconf,amsmath,graphicx,hyperref}
\usepackage{xspace}
\usepackage{physics} 
\usepackage{amsfonts}
\usepackage{booktabs} % 三线表
\usepackage{tabularx} % 自动调整宽度表格
\usepackage{makecell} % 单元格格式控制
\usepackage{multirow} % 多行合并
\usepackage{siunitx} % 数字对齐
\usepackage{caption} % 标题格式
\usepackage{algorithm}
\usepackage{algorithmic}

\usepackage[switch]{lineno}
\newcommand{\framework}{\textsc{fedcompass}\xspace}

\title{Fedcompass: Federated Clustered and Periodic Aggregation Framework for Hybrid Classical-Quantum Models}

\name{Yueheng Wang$^{1}$, Xing He$^{2}$, Zinuo Cai$^{3}$, Rui Zhang$^{1}$, Ruhui Ma$^{3}$, Yuan Liu$^{1}$, Rajkumar Buyya$^{4}$
\thanks{This work was supported by the Natural Science Foundation of Jiangsu Province under Grant BK20241606, and the National Natural Science Foundation of China under Grant 62472200.} \thanks{$^{*}$ Corresponding author: Rui Zhang(zhangruisg111@jiangnan.edu.cn).}}

\address{$^{1}$School of Artificial Intelligence and Computer Science, Jiangnan University \\
$^{2}$Antai College of Economics \& Management, Shanghai Jiao Tong University \\
$^{3}$School of Computer Science, Shanghai Jiao Tong University \\
$^{4}$Cloud Computing and Distributed Systems Laboratory, University of Melbourne
}

\begin{document}
% \ninept
%
\maketitle
\begin{abstract}
Federated learning enables collaborative model training across decentralized clients under privacy constraints. 
% However, real-world client data often exhibits non-IID characteristics, which challenge federated learning. 
Quantum computing offers potential for alleviating computational and communication burdens in federated learning, yet hybrid classical-quantum federated learning remains susceptible to performance degradation under non-IID data. 
To address this, we propose \framework, a layered aggregation framework for hybrid classical-quantum federated learning. \framework employs spectral clustering to group clients by class distribution similarity and performs cluster-wise aggregation for classical feature extractors. For quantum parameters, it uses circular mean aggregation combined with adaptive optimization to ensure stable global updates. Experiments on three benchmark datasets show that \framework improves test accuracy by up to 10.22\% and enhances convergence stability under non-IID settings, outperforming six strong federated learning baselines.
\end{abstract}
\section{Introduction}
\label{sec:intro}

% quantum computing

% QFL

% Significance of Federated Learning
Nowadays, data privacy and potential leakage risks have become critical issues requiring urgent attention. Federated learning (FL)~\cite{zhang2021survey}, as a privacy-preserving distributed learning paradigm, effectively reduces the privacy risk by training models locally on client devices and uploading only parameter updates instead of raw data. Thanks to this characteristic, FL has been widely adopted in scenarios such as mobile healthcare~\cite{antunes2022federated}, IoT~\cite{lim2020federated}, and distributed sensing~\cite{wang2020federated}. However, this privacy protection mechanism also introduces significant communication and computational overhead, posing serious challenges to training efficiency, especially in large-scale or resource-constrained edge environments~\cite{zhang2025qfi}.

% Quantum Computing as a New Paradigm and NISQ-era Hybrid Quantum-Classical Machine Learning
At the same time, as an emerging computational paradigm, quantum computing~\cite{rietsche2022quantum} leverages quantum superposition and entanglement to enable parallel information processing.
% offering new pathways to overcome the performance limitations of classical computing. 
Since current quantum devices remain in the Noisy Intermediate-Scale Quantum (NISQ) era~\cite{lau2022nisq} with limited error correction capabilities and communication reliability issues~\cite{lin2025hwdsqp}, hybrid classical-quantum machine learning~\cite{de2022survey} has shown great potential. Such architecture employs classical neural networks for efficient feature extraction, and leverages quantum circuits to accelerate computation and enhance representational learning in specific tasks~\cite{chen2021end}, thereby opening new avenues for improving federated learning efficiency.

% Quantum Computing’s Promoting Role in Federated Learning

%non-iid
However, federated learning often faces the challenge of non-IID data~\cite{li2022federated} in practice, which can easily lead to local model bias and difficulties in global convergence~\cite{li2019convergence}. Although methods such as FedProx~\cite{li2020federated}, FedBN~\cite{li2021fedbn}, and FedPer~\cite{arivazhagan2019federated} have achieved certain success in traditional scenarios, non-IID data still presents two major challenges in the classical-quantum federated learning setting. Firstly, significant differences in feature distributions among clients exacerbate the deviation in classical feature extraction layers, making it difficult to maintain global consistency after model aggregation. Secondly, quantum parameters are periodic~\cite{schuld2021effect} and sensitive to data distribution~\cite{huang2021power}. Direct arithmetic averaging can easily cause parameter period mismatch, aggravating training instability and leading to optimization conflicts between classical and quantum modules.
% , thereby weakening model convergence and generalization capabilities.

% problem

% solution
To address the aforementioned challenges, we propose \framework, a layered aggregation optimization framework in the hybrid classical-quantum federated learning setting. To mitigate model deviation caused by differences in client feature distributions, \framework employs spectral clustering based on client category statistics and performs weighted aggregation within clusters to generate cluster-level classical feature extractors. To tackle the periodicity issue specific to quantum parameters, we introduce a circular mean for periodic parameters based on the unit circle, combined with an adaptive optimizer to achieve robust global updates.

To validate the effectiveness and generalization capability of the proposed framework, we conduct extensive experiments on three datasets: MNIST, Fashion-MNIST, and CIFAR-10. The results demonstrate that \framework significantly improves test accuracy and convergence stability across various non-IID settings, consistently outperforming six mainstream federated learning baseline methods. The framework fully exemplifies its comprehensive advantages in accuracy, stability, and privacy preservation within hybrid classical-quantum federated learning environments.

\begin{figure}[t]
\begin{minipage}[b]{1.0\linewidth}
  \centering
  \centerline{\includegraphics[width=8.5cm]{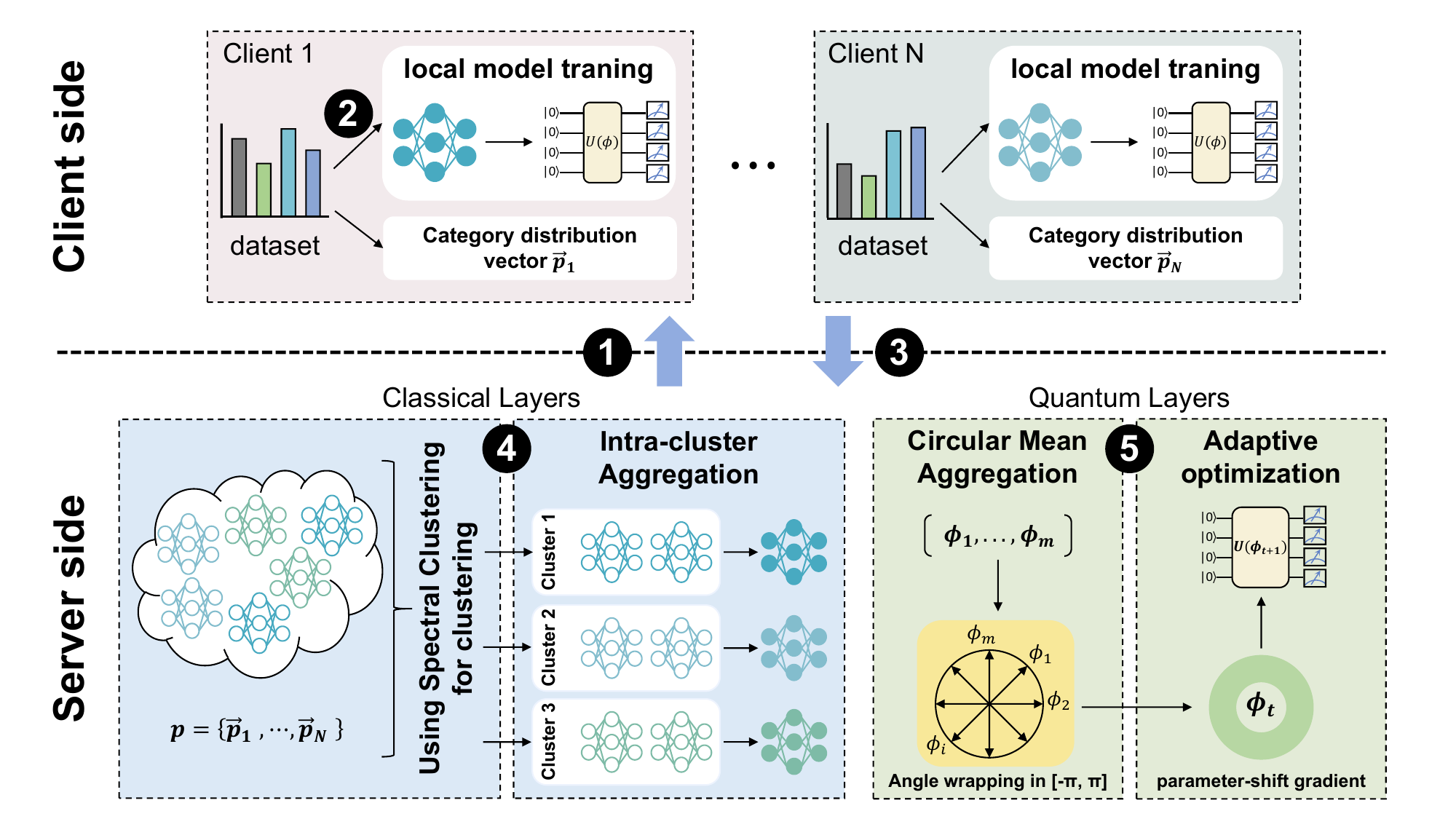}}
%  \vspace{2.0cm}
\end{minipage}
\caption{Overview of our mechanism.}
\label{fig1}
\end{figure}

\section{The Proposed \framework Algorithm}
\label{sec:design}

\subsection{Overview}
Fig.~\ref{fig1} illustrates the overall workflow of \framework. The server first initializes a hybrid global model composed of a classical feature extraction layer and a quantum classifier, which is distributed to all clients (Step 1). Subsequently, each client conducts end-to-end training of the model based on their local data (Step 2). After training, clients upload the updated model parameters along with their data class distribution vectors to the server (Step 3). The server employs a hierarchical aggregation strategy, processing the classical and quantum layers separately. For the classical layer, clients are dynamically clustered based on data distribution similarity, and weighted aggregation is performed within each cluster to generate cluster-level classical feature extractors (Step 4). For the quantum layer, the circular mean method is applied to aggregate quantum parameters, combined with an adaptive optimization strategy to achieve robust global updates (Step 5). Finally, the server distributes the updated cluster-level classical model parameters and global quantum parameters to the clients for the next round of training.

\subsection{Cluster-Based Aggregation for Classical Feature Extraction Layer}

To address model bias caused by non-IID data distributions, we introduce a client clustering mechanism based on data distribution similarity in the classical network part, enhancing the model's adaptability to data heterogeneity. Throughout the federated learning process, no raw data is uploaded, thereby ensuring privacy protection. Let the client set be \( \{c_1, c_2, \ldots, c_N \}\), with corresponding datasets \(\{d_1, d_2, \ldots, d_N \}\), a total number of classes \(C\), and a data concentration parameter \(\alpha\). During data partitioning, the concentration parameter \(\alpha\) controls the degree of data heterogeneity: a smaller \(\alpha\) indicates a more heterogeneous data distribution. Simultaneously, each client uploads its class distribution vector \(\vec{p}\) to the server as a statistical representation of its data features. This vector is a \(C\)-dimensional vector representing the proportion of samples from each class in the client’s local data, i.e., \(\vec{p}_{i}  = ( p_{i1}, p_{i2}, …, p_{iC} )\), where \( p_{ij}\) denotes the proportion of class \(j\) data in client \(c_i\). 

The server collects the local data distribution vectors \(\{{\vec{{p}}_1, \vec{{p}}_2, \ldots, \vec{{p}}_N} \}\) from the clients and computes a similarity matrix \(S\) based on these statistics:
\begin{align}\label{eq1}
    S_{ij} = \exp\left( -\lambda_1 \mathrm{JS}(\vec{{p}}_i,\vec{{p}}_j) - \lambda_2 \frac{\left | n_i - n_j \right |}{ n_i + n_j } \right).
\end{align}

The similarity metric comprehensively evaluates the similarity between clients by considering both distribution divergence and sample size discrepancy. The Jensen–Shannon divergence \(\mathrm{JS}(\cdot,\cdot)\) measures the difference in class distribution patterns between clients. The second term serves as the relative difference in sample size, where \(n_i\) denotes the sample size of client \(c_i\), reflecting the impact of data volume disparity on model updates. The hyperparameters \(\lambda_1\) and \(\lambda_2\) balance the weights of the two terms: increasing \(\lambda_1\) places greater emphasis on distribution consistency, while increasing \(\lambda_2\) prioritizes the alignment of sample size scales.

Based on this, the server employs a spectral clustering algorithm grounded in the Normalized Cut criterion. It computes the normalized Laplacian matrix, performs eigenvalue decomposition, and applies K-means clustering to the top \(M\) eigenvectors to identify groups of clients with similar data distribution patterns. The clients are then grouped into \(M\) clusters \(\{\mathcal{C}_1, \mathcal{C}_2, \ldots, \mathcal{C}_M\}\), with each cluster corresponding to a potential data distribution pattern, thereby achieving an effective partitioning of heterogeneous client groups.

During the aggregation phase, the server receives the classical feature extraction layer parameters \(\theta_c^{(i)}\) uploaded by the clients and performs a weighted average aggregation within each cluster:
\begin{align}\label{eq2}
    \theta_{c}^{(m)} = \frac{ \sum_{i \in \mathcal{C}_m} n_i \cdot \theta_{c}^{(i)} }{ \sum_{i \in \mathcal{C}_m} n_i },
\end{align}
where the weights are determined by the local sample size of each client, resulting in a cluster-shared classical model.

\begin{algorithm}[t]
\caption{Quantum Parameter Update and Aggregation.}
\label{alg:quantum_update}
\begin{algorithmic}[1]
\STATE \textbf{Input:} Client parameters $\{\boldsymbol{\phi}_i\}_{i=1}^N$, sample sizes $\{n_i\}_{i=1}^N$, previous global parameters $\boldsymbol{\phi}_t$, FedAdam states $m_{t-1}, v_{t-1}$, hyperparameters $\beta_1, \beta_2, \eta, \epsilon$
\STATE \textbf{Output:} Updated global parameters $\boldsymbol{\phi}_{t+1}$, updated states $m_t, v_t$
\STATE Compute client weights: $\omega_i = n_i / \sum_j n_j$
\FOR{each dimension $j = 1, \ldots, m$}
    % \STATE $S_j = \sum_i \omega_i \sin(\phi_i^{(j)})$
    % \STATE $C_j = \sum_i \omega_i \cos(\phi_i^{(j)})$
    % \STATE $\bar{\phi}_j = \operatorname{atan2}(S_j, C_j)$
    \STATE Compute $\bar{\phi}_j$ via Eq.~\ref{eq3} \COMMENT{Circular mean aggregation}
\ENDFOR
\STATE Aggregated parameters: $\bar{\boldsymbol{\phi}} = (\bar{\phi}_1, \ldots, \bar{\phi}_m)$
\STATE Construct gradient: $\mathbf{g}_t = \boldsymbol{\phi}_t - \bar{\boldsymbol{\phi}}$
\STATE Update moment: $m_t,v_t$ via Eq.~\ref{eq4}
\STATE Bias correction: $\hat{m}_t,\hat{v}_t$ via Eq.~\ref{eq5}
\STATE Global update: $\boldsymbol{\phi}_{t+1}$ via Eq.~\ref{eq6}
\RETURN $\boldsymbol{\phi}_{t+1}, m_t, v_t$
\end{algorithmic}
\end{algorithm}

\subsection{Global Aggregation Optimization for Periodic Quantum Classifier}
As a globally shared module, the quantum classifier requires consistency and stability in its parameters across all clients. To achieve this, we design a quantum parameter aggregation method on the server side based on periodic averaging and adaptive updating. This approach first employs circular mean to resolve inconsistencies caused by the periodicity of rotation angles, and then introduces an adaptive update mechanism to enhance the stability of global convergence.

The overall quantum parameter aggregation and update process is described in Algorithm \ref{alg:quantum_update}. Let the quantum parameters uploaded by client \(c_i\) be \(\boldsymbol{\phi}_i = (\phi_i^{(1)}, \phi_i^{(2)}, \ldots, \phi_i^{(m)})\). For the \(j\)-th parameter dimension, the aggregation process is defined as:
\begin{align}\label{eq3}
    \bar{\phi}_j = \operatorname{atan2}\left( \sum_{i=1}^{N} \omega_i \sin(\phi_i^{(j)}), \sum_{i=1}^{N} \omega_i \cos(\phi_i^{(j)})\right),
\end{align}
where \(\omega_{i}={n_{i}}/{\sum_{j}n_{j} }\) is the client weight. This operation (line 5)  maps angles to the unit circle for averaging before mapping them back to the angular space, thereby avoiding periodicity-induced inconsistencies.

To further enhance the convergence stability of the global quantum classifier, we employ an adaptive optimizer on the server side (lines 7–11) to update the aggregated parameters globally. First, the average gradient \(\mathbf{g}_t\) for the quantum parameters in round t is computed based on the global quantum parameters \(\mathbf{g}_t = \boldsymbol{\phi}_t - \bar{\boldsymbol{\phi}}\). Subsequently, momentum update, bias correction, and parameter update are performed as follows:
\begin{equation}\label{eq4}
m_t = \beta_1 m_{t-1} + (1-\beta_1)\mathbf{g}_t, \quad 
v_t = \beta_2 v_{t-1} + (1-\beta_2)\mathbf{g}_t^2,
\end{equation}
\begin{equation}\label{eq5}
\hat{m}_t = \frac{m_t}{1-\beta_1^t}, \quad 
\hat{v}_t = \frac{v_t}{1-\beta_2^t},
\end{equation}
\begin{equation}\label{eq6}
\boldsymbol{\phi}_{t+1} = \boldsymbol{\phi}_t 
- \eta \cdot \frac{\hat{m}_t}{\sqrt{\hat{v}_t} + \epsilon},
\end{equation}
where \(\beta_1\) and \(\beta_2\) are momentum hyperparameters, \(\eta\) is the learning rate, and \(\epsilon\) is a numerical stability constant. The integration of the adaptive optimizer with the periodic constraints of quantum parameters ensures stable convergence of the quantum classifier under non-IID data distributions.

Finally, the server distributes the updated cluster-level classical model parameters \(\theta_c^{(m)}\) and the quantum classifier parameters \(\boldsymbol{\phi}_{t+1}\) to the clients for the next round of training and aggregation. This design addresses the periodicity of quantum parameters while facilitating the transfer of discriminative capabilities across different clients through global sharing, thereby improving overall classification performance and convergence stability.

\section{PERFORMANCE EVALUATION}
\label{sec:evaluation}

\subsection{Experiments Setup}

\textbf{Datasets.} We evaluate \framework and comparative methods on three datasets: MNIST, Fashion-MNIST, and CIFAR-10. From each dataset, we uniformly selected 4 classes to form a four-class classification task. The data was partitioned using a Dirichlet distribution~\cite{wang2020federated} with two non-IID parameter settings, \(\alpha=0.3\) and \(\alpha=0.7\).

% \textbf{Models.} We employ a classical-quantum hybrid architecture: the classical network handles feature extraction, and the quantum network performs the classification task. 
% % To reduce the computational and communication burden on the quantum side, we differentially configure the classical-quantum hybrid model based on dataset complexity. 
% For CIFAR-10 and Fashion-MNIST, the first two layers of ResNet-18 are used to extract features, which are then fed into a quantum convolutional network to capture high-dimensional patterns.
% For MNIST, we use LeNet and parameterized quantum circuit, respectively.

\textbf{Models.} We adopt a hybrid classical-quantum architecture, where a classical network performs feature extraction and a quantum network carries out classification. For MNIST, we use LeNet followed by a parameterized quantum circuit. For more complex datasets such as CIFAR-10 and Fashion-MNIST, the first two layers of ResNet-18 are employed for feature extraction, and the features are then passed to a quantum convolutional network.

\textbf{Training Settings.} We simulate a federated learning environment with 10 clients. Each client performs 5 local epochs per communication round with a batch size of 32. Due to the high overhead of quantum training, the server conducts 5 global communication rounds in total. We use the Adam optimizer with a learning rate of 0.001 for updating both the local models and the server-side quantum parameters.

\textbf{Baselines.} We compare \framework with the following six classical federated learning methods: (1) FedAvg~\cite{mcmahan2017communication}; (2) FedProx~\cite{li2020federated}; (3) FedBN~\cite{li2021fedbn}; (4) FedPer~\cite{arivazhagan2019federated}; (5) FedNova~\cite{wang2020tackling}; (6) Scaffold~\cite{karimireddy2020scaffold}.

\textbf{Implementation.} We employ Ray as the distributed computing framework to coordinate multi-client parallel training tasks. The model construction and gradient calculation for the quantum part are implemented using the PennyLane library. The federated learning process is built and executed based on the Flower framework.

\subsection{Results and Discussion}
\begin{figure}[t]
\begin{minipage}[b]{.48\linewidth}
  \centering
  \centerline{\includegraphics[width=4.0cm]{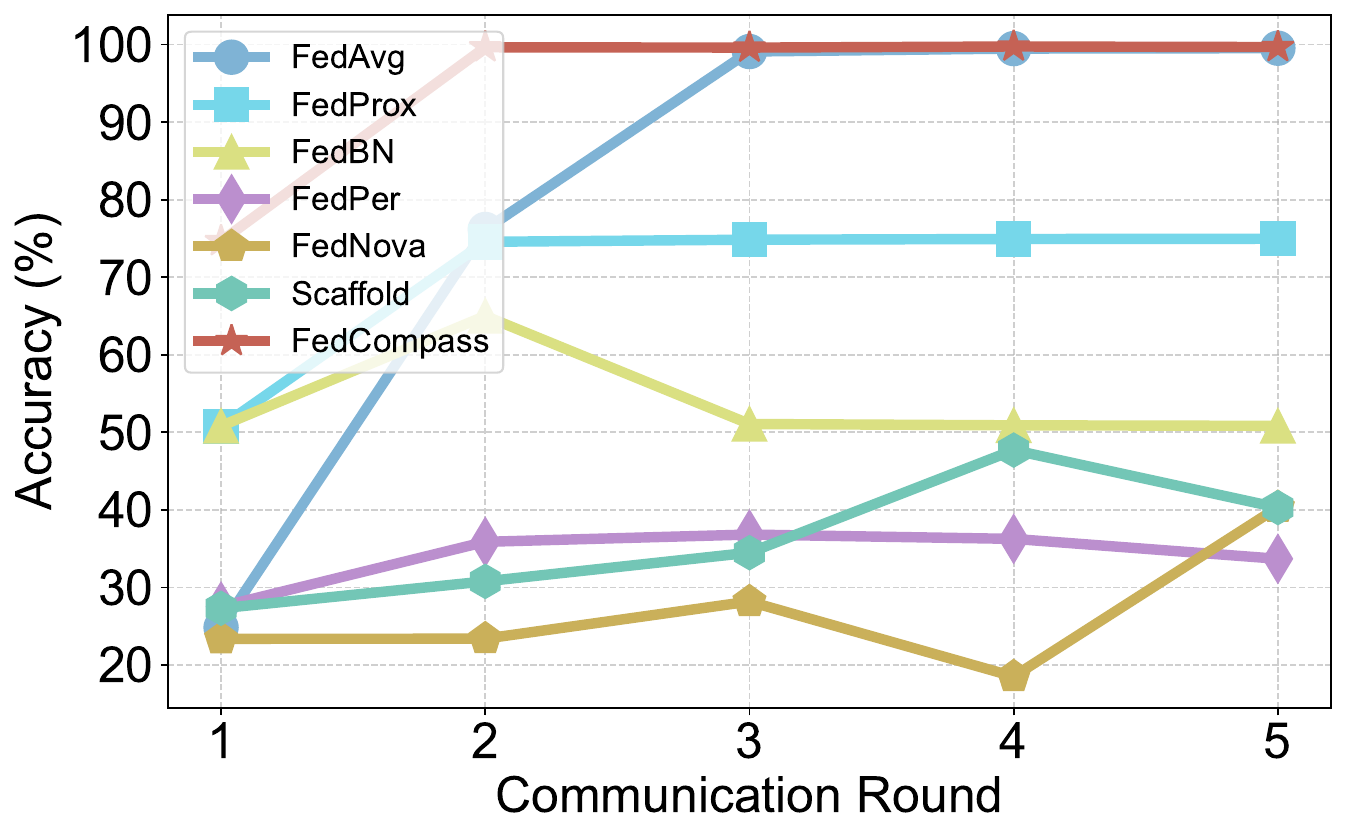}}
%  \vspace{1.5cm}
  \centerline{(a) $\alpha=0.3$}\medskip
\end{minipage}
\hfill
\begin{minipage}[b]{0.48\linewidth}
  \centering
  \centerline{\includegraphics[width=4.0cm]{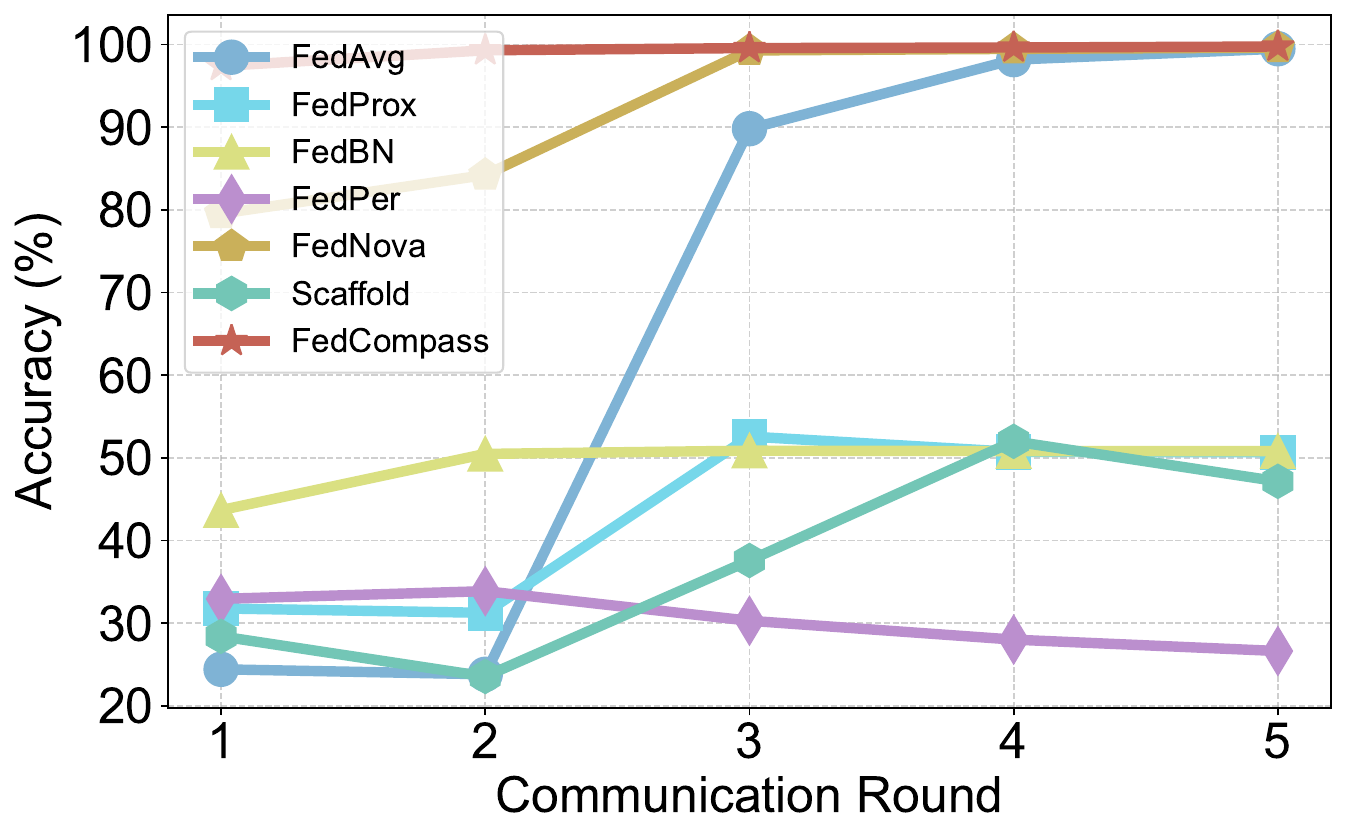}}
%  \vspace{1.5cm}
  \centerline{(b) $\alpha=0.7$}\medskip
\end{minipage}
\caption{Convergence curves of test accuracy versus communication rounds on MNIST under non-IID settings.}
\label{fig:mnist}
\end{figure}

\begin{figure}[t]
\begin{minipage}[b]{.48\linewidth}
  \centering
  \centerline{\includegraphics[width=4.0cm]{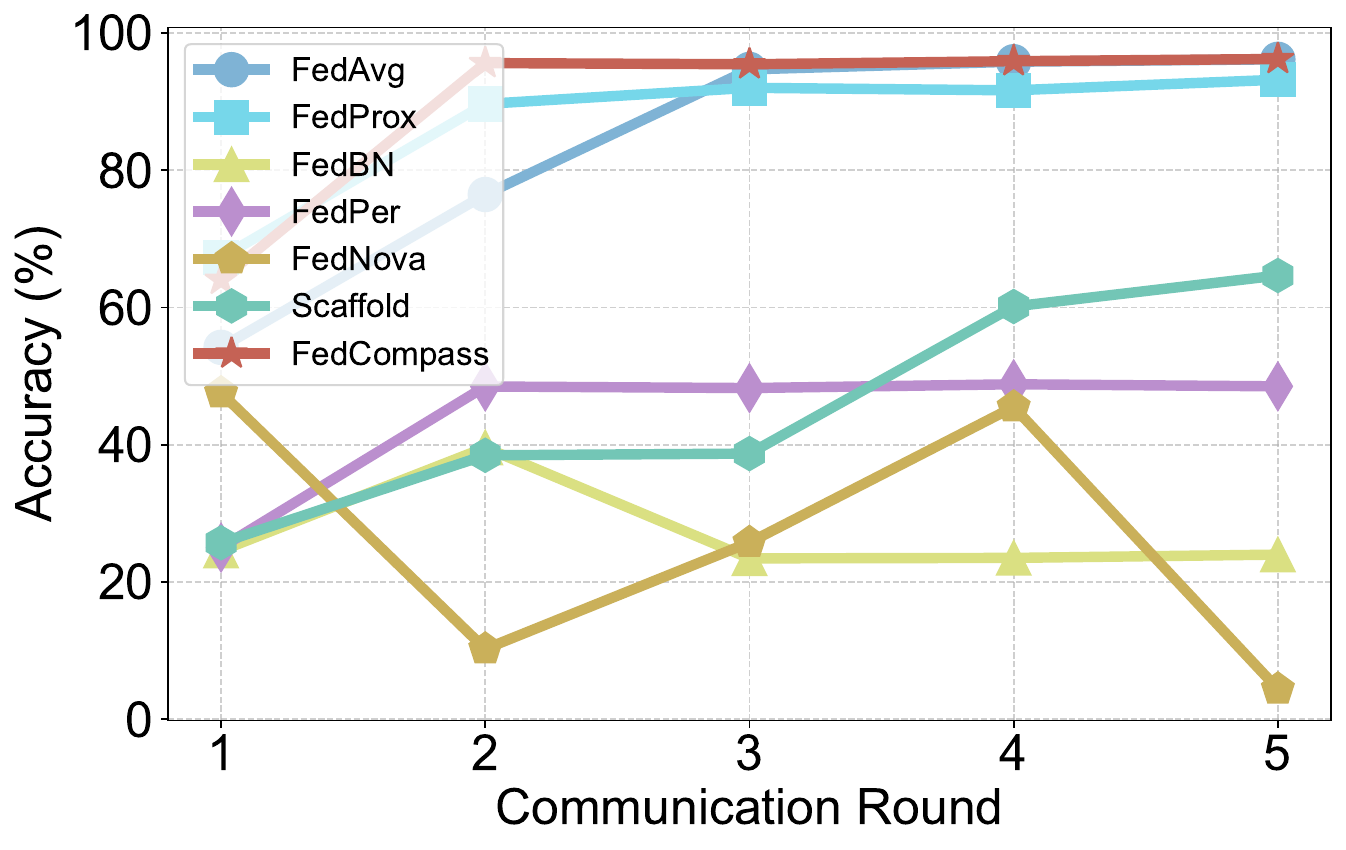}}
%  \vspace{1.5cm}
  \centerline{(a) $\alpha=0.3$}\medskip
\end{minipage}
\hfill
\begin{minipage}[b]{0.48\linewidth}
  \centering
  \centerline{\includegraphics[width=4.0cm]{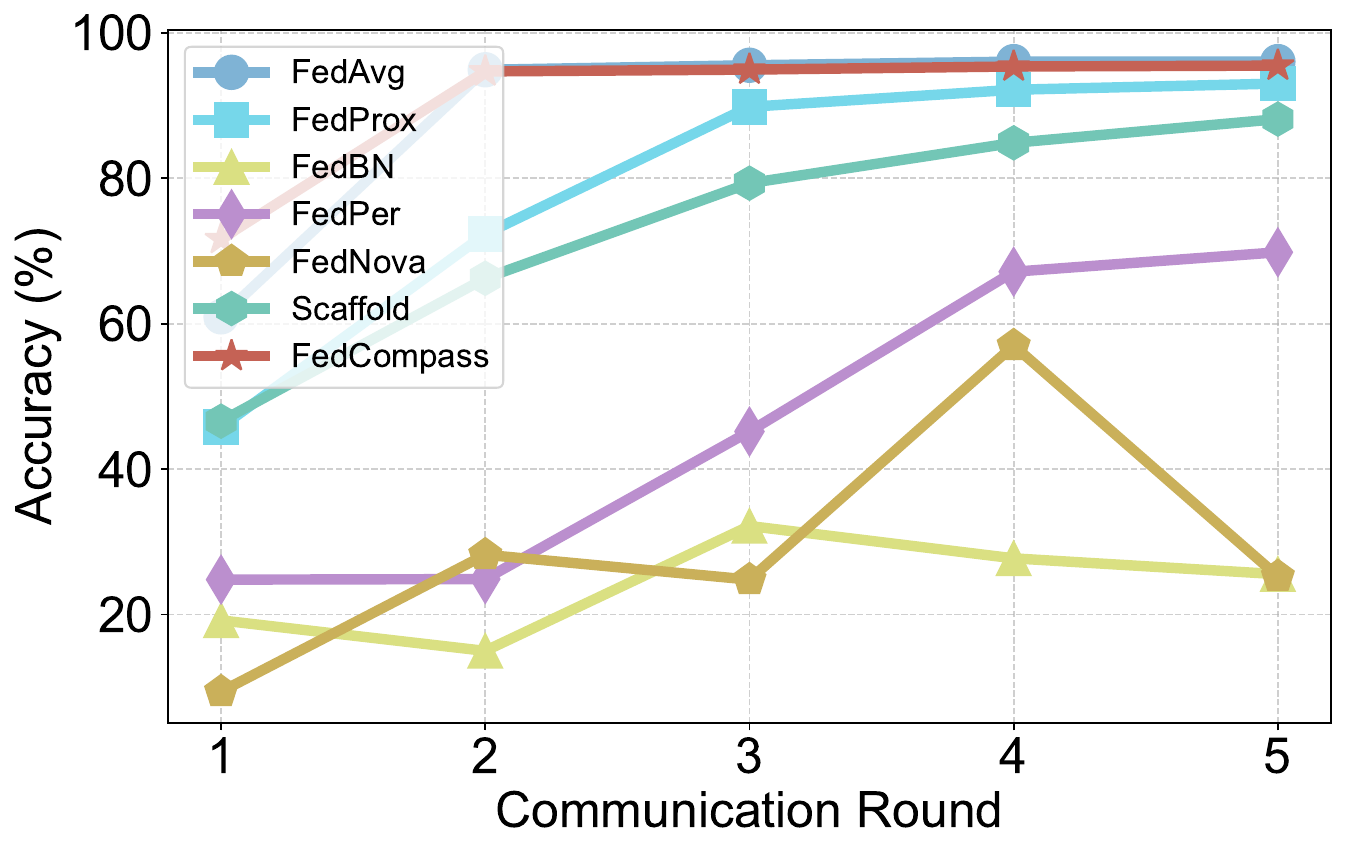}}
%  \vspace{1.5cm}
  \centerline{(b) $\alpha=0.7$}\medskip
\end{minipage}
\caption{Convergence curves of test accuracy versus communication rounds on Fashion-MNIST under non-IID settings.}
\label{fig:fmnist}
\end{figure}

\begin{figure}[!t]
\begin{minipage}[b]{.48\linewidth}
  \centering
  \centerline{\includegraphics[width=4.0cm]{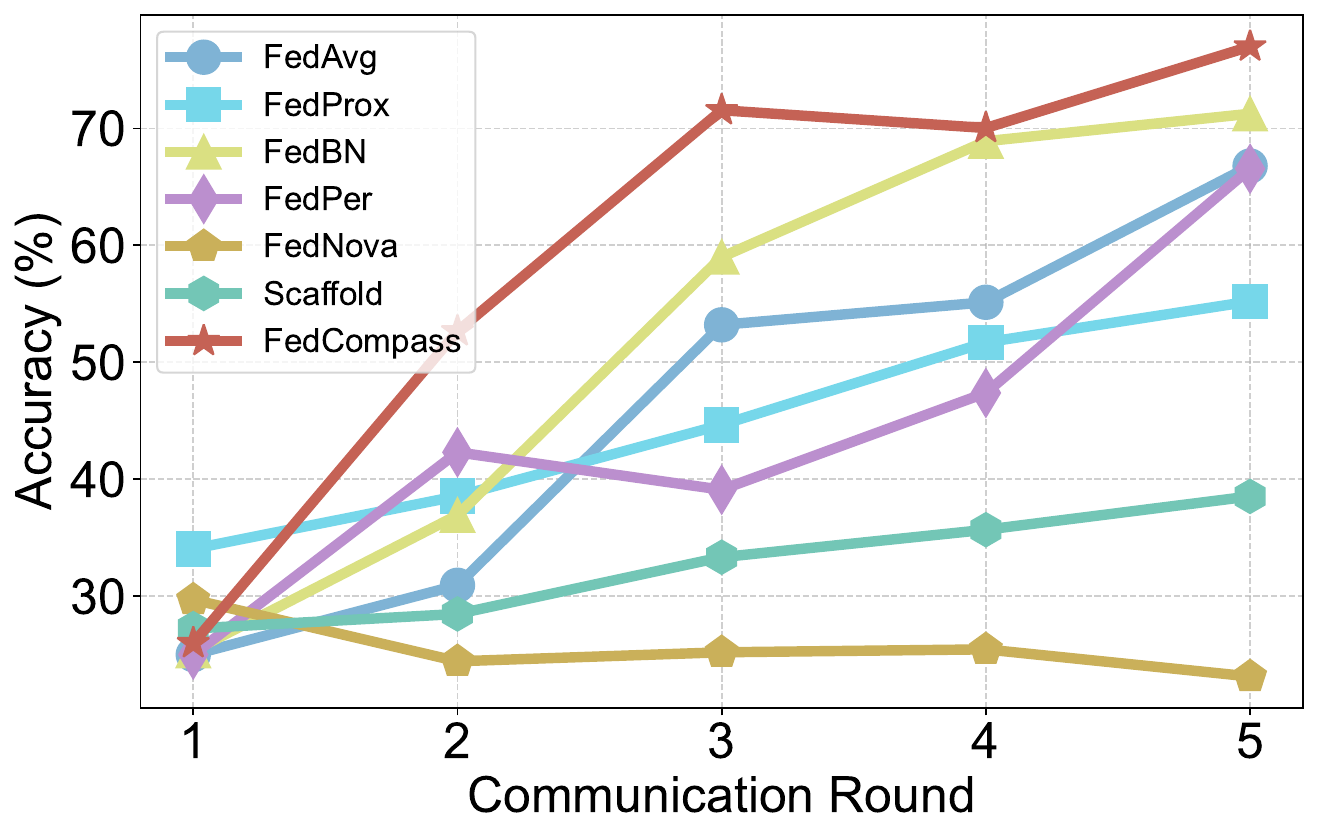}}
%  \vspace{1.5cm}
  \centerline{(a) $\alpha=0.3$}\medskip
\end{minipage}
\hfill
\begin{minipage}[b]{0.48\linewidth}
  \centering
  \centerline{\includegraphics[width=4.0cm]{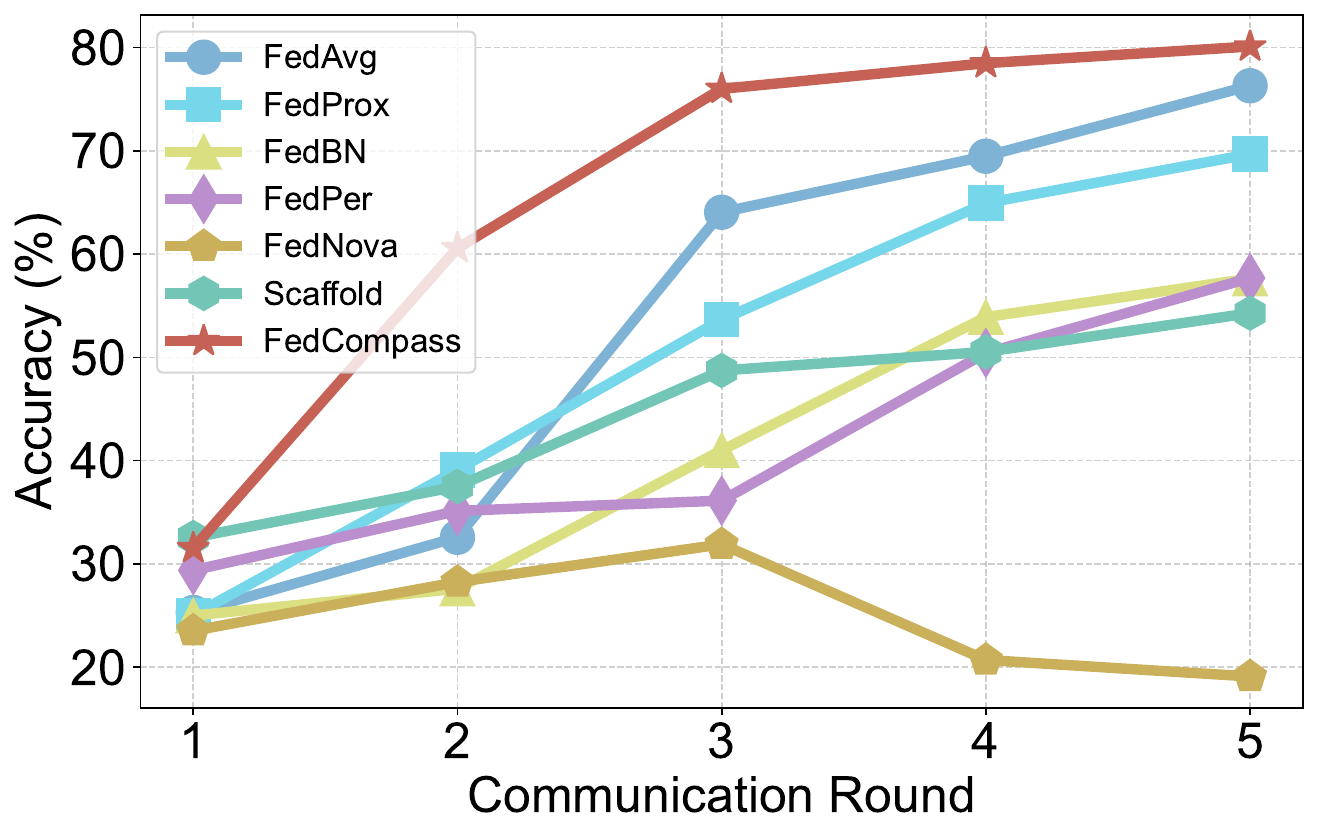}}
%  \vspace{1.5cm}
  \centerline{(b) $\alpha=0.7$}\medskip
\end{minipage}
\caption{Convergence curves of test accuracy versus communication rounds on CIFAR-10 under non-IID settings.}
\label{fig:cifar10}
\end{figure}

We evaluate \framework on MNIST, Fashion-MNIST, and CIFAR-10 under two non-IID settings with \(\alpha\) = 0.3 and \(\alpha\) = 0.7, comparing it against six baseline methods. As shown in the accuracy results (Table~\ref{tab:acc}) and convergence curves (Fig.~\ref{fig:mnist} --~\ref{fig:cifar10}), \framework consistently achieves the best performance in most scenarios.

\begin{table}[t]
\caption{Comparison of test accuracy of different federated learning algorithms across three datasets under non-IID settings. Best values are in \textbf{bold} and second best are \underline{underlined}.}
\label{tab:acc}
\centering
\footnotesize
\setlength{\tabcolsep}{4pt}
\begin{tabular}{@{}ccccccc@{}}
\toprule
Dataset & \multicolumn{2}{c}{MNIST} & \multicolumn{2}{c}{Fashion-MNIST} & \multicolumn{2}{c}{CIFAR-10} \\
\cmidrule(lr){2-3} \cmidrule(lr){4-5} \cmidrule(lr){6-7}
Non-IID Degree & 0.30 & 0.70 & 0.30 & 0.70 & 0.30 & 0.70 \\
\midrule
FedAvg~\cite{mcmahan2017communication} & \underline{99.54} & 99.49 & \underline{96.15} & \textbf{\textcolor{black}{96.05}} & 66.78 & \underline{76.30} \\
FedProx~\cite{li2020federated} & 74.96 & 50.69 & 93.18 & 93.03 & 55.20 & 69.68 \\
FedBN~\cite{li2021fedbn} & 50.83 & 50.83 & 23.98 & 25.55 & \underline{71.25} & 57.63 \\
FedPer~\cite{arivazhagan2019federated} & 33.68 & 26.65 & 48.50 & 69.83 & 66.55 & 57.68 \\
FedNova~\cite{wang2020tackling} & 40.32 & \underline{99.62} & 4.40 & 25.25 & 23.13 & 19.10 \\
Scaffold~\cite{karimireddy2020scaffold} & 40.29 & 47.17 & 64.65 & 85.15 & 38.53 & 54.28 \\
\framework (Ours) & \textbf{\textcolor{black}{99.69}} & \textbf{\textcolor{black}{99.76}} & \textbf{\textcolor{black}{96.20}} & \underline{95.50} & \textbf{\textcolor{black}{77.00}} & \textbf{\textcolor{black}{80.10}} \\
\bottomrule
\end{tabular}
\end{table}

\begin{table}[t]
\centering
\caption{Test accuracy of ablation study on CIFAR-10 across communication rounds.}
\label{tab:cifar10_ablation}
\begin{tabular}{cccc}
\toprule
Round & No Clustering & No Circular Mean & FedCompass \\
\midrule
1     & 25.10         & 25.00            & 25.98      \\
2     & 50.65         & 38.60            & 52.65      \\
3     & 50.38         & 46.23            & 71.55      \\
4     & 55.15         & 32.85            & 70.03      \\
5     & 56.13         & 26.25            & \underline{77.00}      \\
\bottomrule
\end{tabular}
\end{table}

%mnist

%fmnist

% cifar10

\framework demonstrates the most significant improvement on the CIFAR-10 dataset. Under the condition of \(\alpha\) = 0.3, it achieves an accuracy of 77.00\%, which is a 10.22\% increase compared to FedAvg. This result depends on \framework's clustering mechanism, which effectively groups clients with similar class distributions, thereby reducing discrepancies in classical feature learning. When \(\alpha\) = 0.7, the accuracy further improves to 80.10\%, outperforming FedAvg by 3.80\%, confirming the robustness of our method across varying degrees of non-IID data. On MNIST, its performance approaches the dataset's upper limit of 99.7\%, while on Fashion-MNIST, the improvement is relatively smaller, indicating the dataset's lower sensitivity to distribution shifts. Nonetheless, \framework still maintains leading results. Convergence analysis shows that \framework exhibits faster and more stable convergence under different heterogeneity conditions, with the advantage being particularly pronounced at \(\alpha\) = 0.3.

In the ablation study(Table~\ref{tab:cifar10_ablation}), \framework achieved the highest test accuracy, with steady improvement as the number of communication rounds increased, demonstrating the effectiveness of the complete framework. Removing the clustering mechanism for the classical layers resulted in a significant performance drop, particularly in the later rounds. This indicates that the absence of clustering grouping leads to divergence in the feature extractors, adversely affecting global convergence. Omitting the circular mean aggregation for quantum parameters yielded the lowest and most unstable performance. The substantial fluctuations reflect the adverse impact of misaligned periodicity in quantum rotation angles, underscoring the necessity of circular aggregation for coordinating periodic updates and avoiding optimization conflicts.

% \begin{figure}[t]
% \begin{minipage}[b]{1.0\linewidth}
%   \centering
%   \centerline{\includegraphics[width=8.5cm]{ablation study.pdf}}
% %  \vspace{2.0cm}
%   % \centerline{Ablation Study(\%) on CIFAR-10 ($\alpha=0.3$).}\medskip
% \end{minipage}
% \caption{Ablation Study on Convergence Performance over Communication Rounds on CIFAR-10 ($\alpha=0.3$).}
% \label{fig:ablation}
% \end{figure}

\section{Conclusion}
\label{sec:conclusion}
This paper proposes \framework,  a novel hybrid classical-quantum federated learning framework designed to address the challenges of non-IID data distribution. The method implements two key mechanisms, a spectral clustering-based client grouping strategy with within-cluster aggregation of classical feature extractors, and a circular mean aggregation method combined with adaptive optimization tailored for the periodic nature of quantum parameters. It provides an effective solution to data heterogeneity in hybrid federated learning while enhancing overall performance without compromising convergence.
% We will investigate client-side quantum computing for local model training, alongside adaptive clustering, privacy-preserving mechanisms, and large-scale validation on real quantum hardware.

% \section{Acknowledgment}
% This work was supported by the Natural Science Foundation of Jiangsu Province under Grant BK20241606, and the National Natural Science Foundation of China under Grant 62472200. Rui Zhang is the corresponding author.

% \section{Compliance with Ethical Standards}
% This is a numerical simulation study for which no ethical approval was required. 

\bibliographystyle{IEEEbib}
\bibliography{strings,refs}

\end{document}